\documentclass[10pt,twocolumn,letterpaper]{article}

\usepackage{cvpr}              

\usepackage{graphicx}
\usepackage{amsmath}
\usepackage{amssymb}
\usepackage{booktabs}

\usepackage{bm}
\usepackage{amsmath,amssymb}
\usepackage{mathrsfs}
\usepackage{multirow}
\usepackage{subcaption}
\usepackage{bbm}
\usepackage[accsupp]{axessibility}  

\usepackage[pagebackref,breaklinks,colorlinks]{hyperref}

\usepackage[capitalize]{cleveref}
\crefname{section}{Sec.}{Secs.}
\Crefname{section}{Section}{Sections}
\Crefname{table}{Tab.}{Tab.s}
\crefname{table}{Tab.}{Tabs.}

\def\cvprPaperID{9206} 
\def\confName{CVPR}
\def\confYear{2023}

\begin{document}

\title{Focused and Collaborative Feedback Integration\\for Interactive Image Segmentation}

\author{Qiaoqiao Wei\qquad Hui Zhang\qquad Jun-Hai Yong\\
School of Software, BNRist, Tsinghua University, Beijing, China\\
{\tt\small wqq18@mails.tsinghua.edu.cn \qquad \{huizhang,yongjh\}@tsinghua.edu.cn}
}
\maketitle

\begin{abstract}
Interactive image segmentation aims at obtaining a segmentation mask for an image using simple user annotations.
During each round of interaction,
the segmentation result from the previous round serves as feedback to guide the user's annotation
and provides dense prior information for the segmentation model,
effectively acting as a bridge between interactions.
Existing methods overlook the importance of feedback or simply concatenate it with the original input,
leading to underutilization of feedback and an increase in the number of required annotations.
To address this,
we propose an approach called Focused and Collaborative Feedback Integration (FCFI) to fully exploit the feedback for click-based interactive image segmentation.
FCFI first focuses on a local area around the new click
and corrects the feedback based on the similarities of high-level features.
It then alternately and collaboratively updates the feedback and deep features
to integrate the feedback into the features.
The efficacy and efficiency of FCFI were validated on four benchmarks,
namely GrabCut, Berkeley, SBD, and DAVIS.
Experimental results show that FCFI achieved new state-of-the-art performance
with less computational overhead than previous methods.
The source code is available at
\url{https://github.com/veizgyauzgyauz/FCFI}.
\end{abstract}

\begin{figure}[t]
  \centering
  \includegraphics[width=1.0\columnwidth]{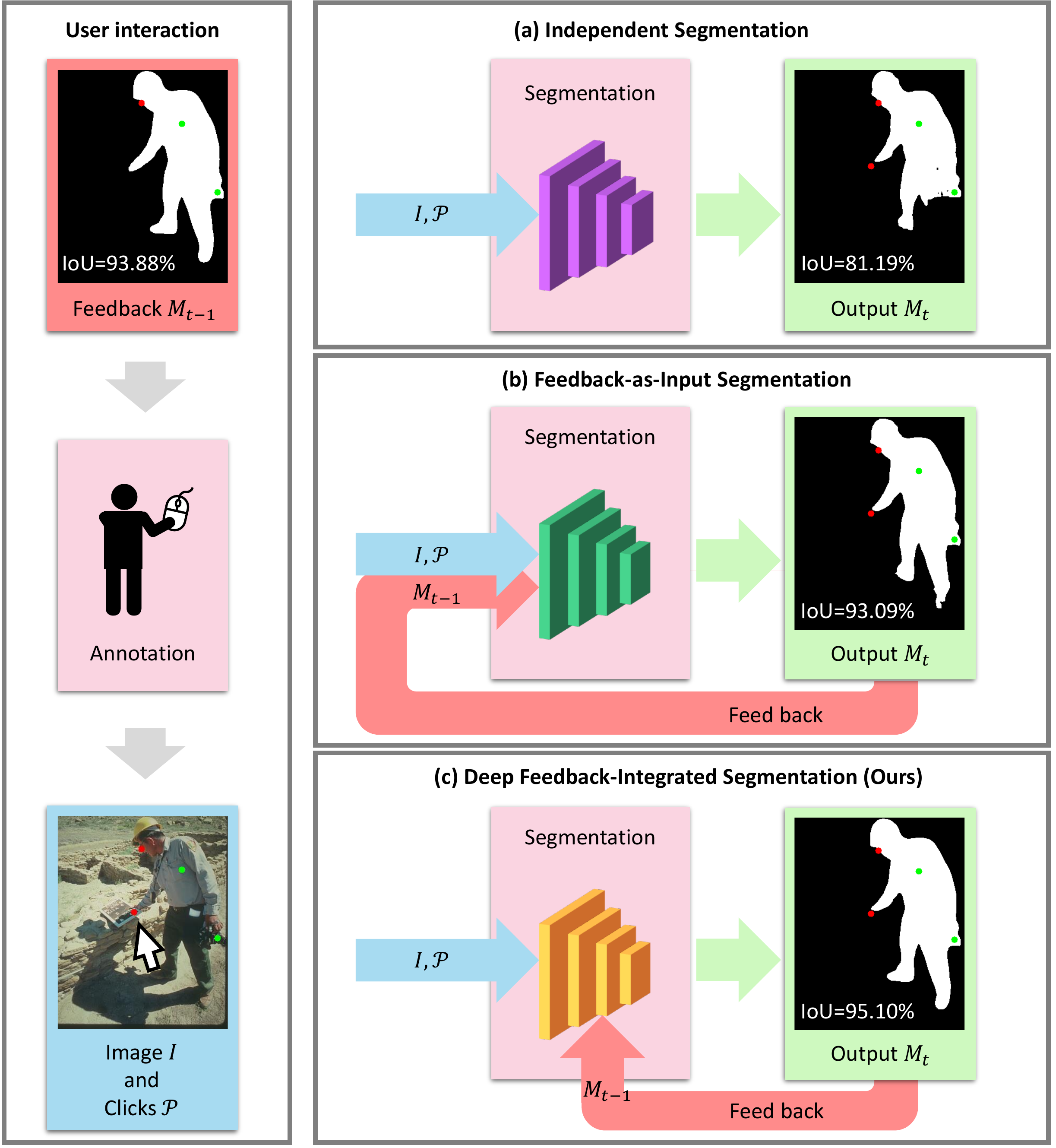}
  \caption{An overview of the interactive system
  and the comparison among (a) independent segmentation \cite{lin2020interactive},
  (b) feedback-as-input segmentation \cite{sofiiuk2021reviving},
  and (c) deep feedback-integrated segmentation.
  See the description in Sec. \ref{sec:introduction}.
  Throughout this paper, green/red clicks represent foreground/background annotations.}
  \label{fig:start}
\end{figure}

\section{Introduction}
\label{sec:introduction}
Interactive image segmentation aims to segment a target object in an image given simple annotations,
such as bounding boxes \cite{rother2004grabcut,lempitsky2009image,wu2014milcut,xu2017deep,yu2017loosecut},
scribbles \cite{bai2014error,li2004lazy,grady2006random},
extreme points \cite{papadopoulos2017extreme,maninis2018deep,benenson2019large,zhang2020interactive},
and clicks \cite{xu2016deep,jang2019interactive,sofiiuk2020fbrs,lin2020interactive,chen2022focalclick,lin2022focuscut}.
Due to its inherent characteristic, i.e., interactivity,
it allows users to add annotations and receive refined segmentation results iteratively.
Unlike semantic segmentation, interactive image segmentation can be applied to unseen categories (categories that do not exist in the training dataset),
demonstrating its generalization ability.
Additionally, compared with instance segmentation,
interactive segmentation is specific since it only localizes the annotated instance.
Owing to these advantages,
interactive image segmentation is a preliminary step for many applications,
including image editing and image composition.
This paper specifically focuses on interactive image segmentation using click annotations
because clicks are less labor-intensive to obtain than other types of annotations.

The process of interaction is illustrated in Fig. \ref{fig:start}.
Given an image, users start by providing one or more clicks to label background or foreground pixels.
The segmentation model then generates an initial segmentation mask for the target object
based on the image and the clicks.
If the segmentation result is unsatisfactory,
users can continue to interact with the segmentation system
by marking new clicks to indicate wrongly segmented regions and obtain a new segmentation mask.
From the second round of interaction,
the segmentation result of the previous interaction - referred to as \textit{feedback} in this paper - will be fed back into the current interaction.
This feedback is instrumental in user labeling and provides the segmentation model with prior information,
which can facilitate convergence and improve the accuracy of segmentation \cite{sofiiuk2021reviving}.


Previous methods have made tremendous progress in interactive image segmentation.
Some of them, such as \cite{majumder2019content,lin2020interactive,sofiiuk2021reviving},
focused on finding useful ways to encode annotated clicks,
while others, like \cite{li2018interactive,jang2019interactive,liew2019multiseg,sofiiuk2020fbrs,chen2021conditional},
explored efficient neural network architectures to fully utilize user annotations.
However, few methods have investigated how to exploit informative segmentation feedback.
Existing methods typically treated each round of interaction independent
\cite{li2018interactive,jang2019interactive,majumder2019content,lin2020interactive,chen2021conditional,hao2021edgeflow}
or simply concatenated feedback with the initial input
\cite{mahadevan2018iteratively,sofiiuk2020fbrs,sofiiuk2021reviving,chen2022focalclick,lin2022focuscut}.
The former approach (Fig. \ref{fig:start}(a)) failed to leverage the prior information provided by feedback,
resulting in a lack of consistency in the segmentation results generated by two adjacent interactions.
In the latter case (Fig. \ref{fig:start}(b)),
feedback was only directly visible to the first layer of the network,
and thus the specific spatial and semantic information it carried would be easily diluted or even lost
through many convolutional layers,
similar to the case of annotated clicks \cite{hao2021edgeflow}.

In this paper, we present Focused and Collaborative Feedback Integration (FCFI)
to exploit the segmentation feedback for click-based interactive image segmentation.
FCFI consists of two modules:
a Focused Feedback Correction Module (FFCM) for local feedback correction
and a Collaborative Feedback Fusion Module (CFFM) for global feedback integration into deep features.
Specifically, 
the FFCM focuses on a local region centered on the new annotated click
to correct feedback.
It measures the feature similarities between each pixel in the region and the click.
The similarities are used as weights to blend the feedback and the annotated label.
The CFFM adopts a collaborative calibration mechanism to integrate the feedback into deep layers (Fig. \ref{fig:start}(c)).
First, it employs deep features to globally update the corrected feedback for further improving the quality of the feedback.
Then, it fuses the feedback with deep features via a gated residual connection.
Embedded with FCFI,
the segmentation network leveraged the prior information provided by the feedback
and outperformed many previous methods.

\section{Related Work}
\textbf{The Method changes in interactive segmentation.}
Early work approached interactive segmentation based on graph models,
such as graph cuts \cite{boykov2001interactive,rother2004grabcut}
and random walks \cite{grady2006random,dong2015sub}.
These methods build graph models on the low-level features of input images.
Therefore, they are sensitive to user annotations and lack high-level semantic information.
Xu et al. \cite{xu2016deep} first introduced deep learning into interactive image segmentation.
Deep convolutional neural networks are optimized over large amounts of data,
which contributes to robustness and generalization.
Later deep-learning-based methods \cite{li2018interactive,jang2019interactive,chen2021conditional} showed striking improvements.

\textbf{The Local refinement for interactive segmentation.}
To refine the primitive predictions,
previous methods \cite{liew2017regional,chen2021conditional,hao2021edgeflow,chen2022focalclick}
introduced additional convolutional architectures to fulfill the task.
These methods followed a coarse-to-fine scheme
and performed global refinement.
Later, Sofiiuk et al. \cite{sofiiuk2020fbrs} proposed the Zoom-In strategy.
From the third interaction, it cropped and segmented the area within an expanded bounding box of the inferred object.
Recent approaches \cite{chen2022focalclick,lin2022focuscut} focused on local refinement.
After a forward pass of the segmentation network,
they found the largest connected component on the difference between the current prediction and the previous prediction.
Then, they iteratively refined the prediction in the region using the same network or an extra lighter network.
The proposed FFCM differs from these methods in that
it does not need to perform feedforward propagation multiple times.
Instead, it utilizes the features already generated by the backbone of the network
to locally refine the feedback once.
Thus, it is non-parametric and fast.

\textbf{The Exploitation of segmentation feedback.}
Prior work has found that
segmentation feedback has instructive effects on the learning of neural networks.
Mahadevan et al. \cite{mahadevan2018iteratively} were the first to incorporate
feedback from a previous iteration into the segmentation network.
They took the feedback as an optional channel of the input.
Later methods \cite{sofiiuk2020fbrs,sofiiuk2021reviving,chen2022focalclick,lin2022focuscut} followed this
and concatenated the feedback with the input.
However, this naive operation may be suboptimal for feedback integration due to the dilution problem \cite{hao2021edgeflow}.
Different from previous approaches,
the proposed CFFM integrates feedback into deep features.
Since both the feedback and the high-level features contain semantic knowledge of specific objects,
fusing them together helps to improve the segmentation results.


\begin{figure*}[t]
    \centering
    \includegraphics[width=1.0\linewidth]{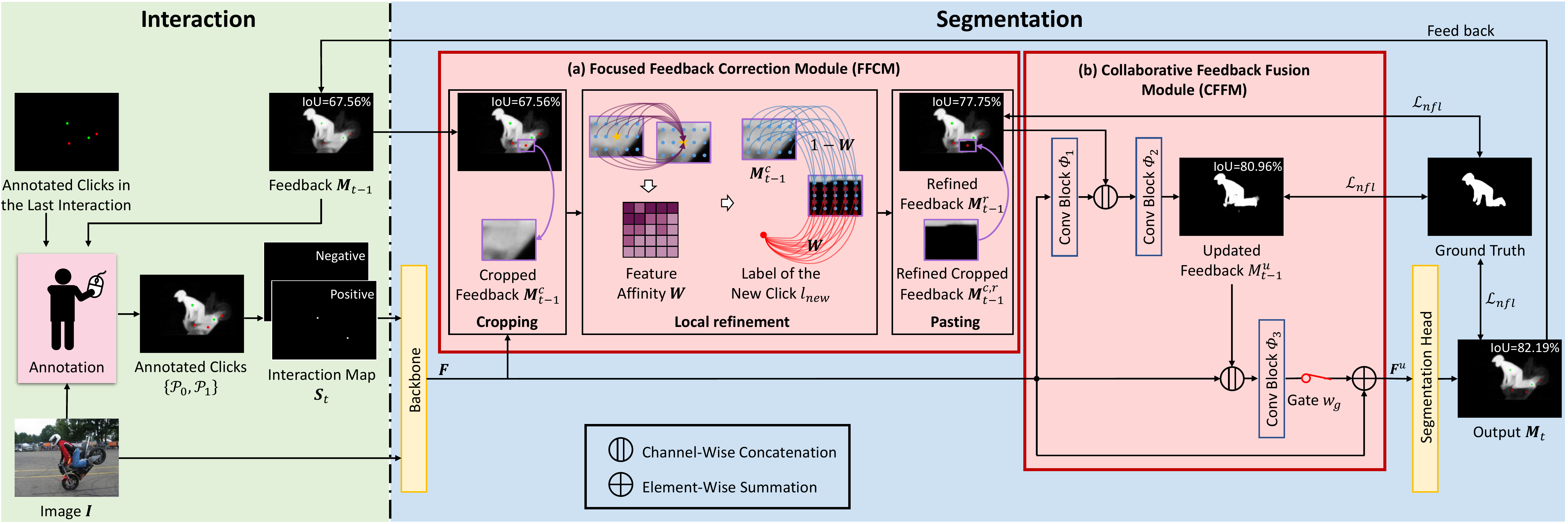}
    \caption{The pipeline of the proposed method. See detailed description in Sec. \ref{sec:method_pipeline}.}
   
    \label{fig:pipeline}
\end{figure*}

\section{Method}
\subsection{Interaction Pipeline}
\label{sec:method_pipeline}
The pipeline of our method is illustrated in Fig. \ref{fig:pipeline}.
The process includes two parts: interaction and segmentation.

\textbf{Interaction.}
In the interaction step,
users give hints about the object of interest
to the segmentation model by providing clicks.
In the first interaction,
users provide clicks
only based on the input image $\bm{I}\in\mathbb{R}^{H\times W\times 3}$,
where $H$ and $W$ denote the height and width, respectively;
in the subsequent interactions,
users mark more clicks according to the input image and the segmentation feedback.
The annotated clicks over all interactions are denoted as $\{\mathcal{P}_0, \mathcal{P}_1\}$,
where $\mathcal{P}_0$ and $\mathcal{P}_1$ represent a background and foreground click set, respectively,
and each element in $\mathcal{P}_0$ and $\mathcal{P}_1$ is the $xy$ position of a click.
The annotated clicks are converted into an interaction map $\bm{S}\in\mathbb{R}^{H\times W\times 2}$ using disk encoding \cite{benenson2019large},
which represents each click as a disk with a small radius.

\textbf{Segmentation.}
In the segmentation step,
a commonly used fully convolutional baseline architecture embedded with FCFI is utilized as the segmentation network.
The baseline network comprises a backbone and a segmentation head,
as shown in the yellow blocks in Fig. \ref{fig:pipeline}.
The backbone gradually captures long-range dependencies of pixels to generate high-level features,
and the segmentation head, including a Sigmoid function, recovers spatial resolutions and extracts semantic information.
To correct feedback locally and incorporate it into the segmentation network,
an FFCM and a CFFM are inserted between the backbone and the segmentation head.
In the $t$-th interaction,
the input image $\bm{I}$ and the interaction map $\bm{S}_t$ are concatenated in depth and fed into the backbone,
and the feedback $\bm{M}_{t-1}$ is directly fed into the FFCM.
The segmentation network outputs a segmentation mask $\bm{M}_{t}\in\mathbb{R}^{H\times W}$.

\subsection{Focused Feedback Correction Module}
\label{sec:FFCM}
From the second interaction,
users are expected to place new clicks on the mislabeled regions.
However, most previous methods \cite{jang2019interactive,majumder2019content,jang2019interactive,sofiiuk2020fbrs,chen2021conditional,sofiiuk2021reviving}
treated all clicks equally and only performed global prediction,
which may weaken the guidance effect of the newly annotated clicks
and cause unexpected changes in correctly segmented regions that are far away from the newly annotated click \cite{chen2022focalclick}.
For instance, in Fig. \ref{fig:start}(a) and (b), comparing $\bm{M}_{t-1}$ and $\bm{M}_t$,
adding a new negative click near the person's right hand modified the segmentation result in the person's feet.
Considering that clicks are usually marked to correct small regions of a segmentation mask,
locally refining the feedback can not only preserve the segmentation results in other regions
but also reduce processing time.
Therefore, we propose the FFCM to correct the feedback from a local view.

As shown in Fig. \ref{fig:pipeline}(a),
the feedback modification in the FFCM requires three steps:
cropping, local refinement, and pasting.
We first narrow down the refinement region,
then refine the feedback using the pixel similarities in the high-level feature space,
and finally fuse the refined feedback with the original one.

\textbf{Cropping.}
To exclude irrelevant regions and focus on the area around the newly annotated click,
we select a small rectangle patch centered on the new annotated click
and crop it from the features $\bm{F}\in\mathbb{R}^{H'\times W'\times 3}$ generated by the backbone,
where $H'$ and $W'$ are the height and width of $\bm{F}$.
The patch has a size of $rH'\times rW'$,
where the expansion ratio $r$ is $0.3$ by default.
The feedback $\bm{M}_{t-1}$
is cropped in the same way.
The cropped features and the cropped feedback are denoted as $\bm{F}^c$ and $\bm{M}_{t-1}^c$, respectively.

\textbf{Local refinement.}
In this step, per-pixel matching is performed.
We measure the feature affinity between each pixel in the patch and the new annotated click
and then blend the feedback with the annotated label of the new click according to the affinity.
The feature affinity $\bm{W}\in\mathbb{R}^{rH'\times rW'}$ is defined as the cosine similarity:
\begin{equation}
  \bm{W}(i,j)=\frac{\bm{F}^c(i,j)\bm{F}^c(x_{new},y_{new})}{||\bm{F}^c(i,j)||_2||\bm{F}^c(x_{new},y_{new})||_2},
\end{equation}
where $(x_{new}, y_{new})$ is the coordinate of the new annotated click in the patch.
Each element in the feature affinity $\bm{W}$ is between 0 to 1.
The larger the affinity is, the more likely that the pixel belongs to the same class (background/foreground) as the new annotated click.
The refined cropped feedback $\bm{M}_{t-1}^r$ is generated
by blending the original feedback $\bm{M}^c_{t-1}$ and the annotated label of the new click $l_{new}$
via a linear combination:
\begin{equation}
  \bm{M}^{c,r}_{t-1}=l_{new}\cdot \bm{W}+(1-\bm{W})\circ\bm{M}^c_{t-1}.
\end{equation}

\textbf{Pasting.}
After obtaining the refined cropped feedback,
we paste it back to the original position on the feedback $\bm{M}_{t-1}$
and denote the refined full-size feedback as $\bm{M}^r_{t-1}$.

To enable optimization in backward-propagation during training,
the cropping and pasting operations are not applied in the FFCM.
Instead, a box mask $\bm{M}_{box}\in\mathbb{R}^{H'\times W'}$
is used to filter out the pixels that are far away from the new annotated click.
A local region is selected first.
It is centered on the new annotated click and has a size of $rH'\times rW'$.
Pixels within the region
are set to 1,
and others are set to 0.
Similar to the local refinement,
we perform global refinement on the full-size features and feedback.
A full-size refined feedback $\bm{M}^{r'}_{t-1}$ is obtained.
Finally, we keep the focused area and obtain the corrected feedback $\bm{M}^r_{t-1}$ as
\begin{equation}
  \bm{M}^r_{t-1}=\bm{M}_{box}\circ\bm{M}^{r'}_{t-1}+(1-\bm{M}_{box})\circ\bm{M}_{t-1}.
\end{equation}


\subsection{Collaborative Feedback Fusion Module}
\label{sec:CFFM}


In an interaction, a user usually adds new clicks based on the segmentation feedback.
Basically, interactions are successive,
and previous segmentation results provide prior information about the location
and shape of the target object for the current interaction.
Therefore, it is natural to integrate feedback into the segmentation network,
and this integration is supposed to improve the segmentation quality.

Previous methods \cite{sofiiuk2020fbrs,sofiiuk2021reviving,chen2022focalclick}
simply concatenated the feedback with the input image and the interaction map
and then fed them into the segmentation network.
However, experimental results show that
this naive method cannot exploit the prior information provided by the feedback.
There are two reasons.
First, early fusion - fusing at the beginning or shallow layers of a network -
easily causes information dilution or loss \cite{hao2021edgeflow}.
Second, from the semantic information perspective,
the feedback contains dense semantic information
compared with the low-level input image and sparse clicks.
Thus, theoretically, fusing the feedback into deep features rather than the input
enables the segmentation network to obtain segmentation priors
while preserving the extracted semantic information.

The CFFM is introduced in this paper to integrate segmentation feedback into high-level features
at deep layers of the segmentation network.
Fig. \ref{fig:pipeline}(b) illustrates the architecture of the CFFM.
The CFFM comprises two information pathways,
called feedback pathway and feature pathway, respectively.
The two pathways utilize the feedback and the high-level features
to collaboratively update each other.

\textbf{Feedback pathway.}
This pathway performs global refinement on the feedback
with the aid of deep features.
First, after being encoded by a convolution block $\Phi_1$,
the features $\bm{F}$
are concatenated with the corrected feedback signal $\bm{M}^r_{t-1}$ in the channel dimension.
Then, convolutional layers followed by a Sigmoid function, denoted as $\Phi_2$,
are applied to update the feedback:
\begin{equation}
  \bm{M}^u_{t-1}=\Phi_2(\text{concat}(\Phi_1(\bm{F};\theta_1),\bm{M}^r_{t-1});\theta_2),
\end{equation}
where $\theta_1$ and $\theta_2$ denote the learnable parameters of $\Phi_1$ and $\Phi_2$,
and ``$\text{concat}(\cdot,\cdot)$'' denotes channel-wise concatenation.

\textbf{Feature pathway.}
In this pathway,
we update the high-level features with the feedback as a guide.
The features are fused with the updated feedback $\bm{M}^u_{t-1}$ through a convolution block $\Phi_3$.
To avoid negative outcomes from useless learned features,
we update the features $\bm{F}$ via a skip connection following ResNet \cite{he2016deep}.
However, the skip connection leads to an inaccurate prediction of the final output $\bm{M}_t$ in the first interaction,
but not in the other interactions.
The reason is that the feedback
is initialized to an all-zero map
and has the maximum information entropy in the first interaction \cite{mahadevan2018iteratively}.
Consequently, it may add noise to deep features.
The reliability of feedback grows dramatically from the first interaction to the second interaction,
making the feedback more instructive.
To tackle this problem,
we introduce a gate $w_g$ to control the information flow from the feedback pathway.
The gate is set to 0 in the first interaction
to prevent the inflow of noise from hurting performance,
and it is set to 1 in the subsequent interactions.
Mathematically, the fused features $\bm{F}^u$ can be formulated as
\begin{equation}
  \label{eq:update_feature}
  \bm{F}^u=w_g\cdot \Phi_3(\text{concat}(\bm{F},\bm{M}^u_{t-1});\theta_3)+\bm{F},
\end{equation}
where $\theta_3$ denotes the learnable parameters of $\Phi_3$.

\subsection{Training Loss}
The normalized focal loss $L_{nfl}$ \cite{sofiiuk2019adaptis} is applied as the objective function
because training with it yields better performance in interactive image segmentation,
which is demonstrated in RITM \cite{sofiiuk2021reviving}.
The constraint is imposed on $\bm{M}^r_{t-1}$, $\bm{M}_t$, and $\bm{M}^u_{t-1}$.
Particularly, we only calculate the loss in the cropped area for $\bm{M}^r_{t-1}$.
The full loss function is
\begin{equation}
  \begin{aligned}
  L=L_{nfl}(\bm{M}^r_{t-1},\bm{M}_{gt})+L_{nfl}(\bm{M}^u_{t-1},\bm{M}_{gt}) \\
  +L_{nfl}(\bm{M}_{t},\bm{M}_{gt}),
  \end{aligned}
\end{equation}
where $\bm{M}_{gt}$ denotes the ground truth segmentation mask.

\begin{table*}[t]
  \begin{center}
  \resizebox{0.99\linewidth}{!}{
  \begin{tabular}{l|l|l|c|c|c|c|c|c|c}
  \toprule
      \multirow{2}{*}{Method} & \multirow{2}{*}{Backbone} & \multirow{2}{*}{Train set} & \multicolumn{2}{c|}{GrabCut} & Berkeley & \multicolumn{2}{c|}{SBD} & \multicolumn{2}{c}{DAVIS} \\
      \cline{4-10}
      &&& NoC@85\% & NoC@90\% & NoC@90\% & NoC@85\% & NoC@90\% & NoC@85\% & NoC@90\% \\
      \midrule
      DOS w/o GC \cite{xu2016deep}           & FCN-8s  & SBD & 8.02 & 12.59 & - & 14.30 & 16.79 & 12.52 & 17.11 \\
      DOS w/ GC \cite{xu2016deep}            & FCN-8s  & SBD & 5.08 & 6.08 & - & 9.22 & 12.80 & 9.03 & 12.58 \\
      RIS-Net \cite{liew2017regional}        & VGG-16  & Pascal VOC  &   -  & 5.00 & - & 6.03 & - &   -  &   -    \\
      LD \cite{li2018interactive}            & VGG-19  & SBD   & 3.20 & 4.79 &   -   & 7.41 &   -  & 5.95 & 9.57 \\
      CAG \cite{majumder2019content}         & FCN-8s & Augmented SBD &   -  & 3.58 & 5.60 & - &   -  &   -  &   -  \\
      BRS \cite{jang2019interactive}         & DenseNet & SBD  & 2.60 & 3.60 & 5.08 & 6.59 & 9.78 & 5.58 & 8.24 \\
      \hline
      f-BRS-B \cite{sofiiuk2020fbrs}         & ResNet-101 & SBD & 2.30 & 2.72 & 4.57 & 4.81 & 7.73 & 5.04 & 7.41 \\
      FCA-Net \cite{lin2020interactive}      & ResNet-101 & Augmented SBD & 1.88 & 2.14 & 4.19 & - & - & 5.38 & 7.90 \\
      IA+SA \cite{kontogianni2020continuous} & ResNet-101 & Augmented SBD & - & 3.07 & 4.94 & - & - & 5.16 & - \\
      CDNet \cite{chen2021conditional}       & ResNet-101 & SBD & 2.42 & 2.76 & 3.65 & 4.73 & 7.66 & 5.33 & 6.97 \\
      FocusCut \cite{lin2022focuscut}        & ResNet-101 & SBD & \textbf{1.46} & \textbf{1.64} & \underline{3.01} & \underline{3.40} & \textbf{5.31} & \underline{4.85} & \textbf{6.22} \\
      \textbf{Ours}                          & ResNet-101 & SBD & \underline{1.64} & \underline{1.80} & \textbf{2.84} & \textbf{3.26} & \underline{5.35} & \textbf{4.75} & \underline{6.48} \\ 
      \hline
      RITM \cite{sofiiuk2021reviving}        & HRNet-18s  & COCO+LVIS & 1.54 & 1.68 & 2.60 & \underline{4.26} & 6.86 & 4.79 & 6.00 \\
      FocalClick-S1 \cite{chen2022focalclick} & HRNet-18s & COCO+LVIS & 1.72 & 1.94 & 3.40 & 4.75 & 7.22 & 5.19 & 7.95 \\
      FocalClick-S2 \cite{chen2022focalclick} & HRNet-18s & COCO+LVIS & \underline{1.52} & \underline{1.66} & \underline{2.41} & 4.37 & \underline{6.59} & \underline{4.20} & \underline{5.49} \\
      \textbf{Ours}                          & HRNet-18s  & COCO+LVIS & \textbf{1.50} & \textbf{1.56} & \textbf{2.05} & \textbf{3.88} & \textbf{6.24} & \textbf{3.70} & \textbf{5.16} \\
      \hline
      EdgeFlow \cite{hao2021edgeflow}        & HRNet-18   & COCO+LVIS & 1.60 & 1.72 & 2.40 & - &   -  & 4.54 & 5.77 \\
      RITM \cite{sofiiuk2021reviving}        & HRNet-18   & COCO+LVIS & \underline{1.42} & \underline{1.54} & \underline{2.26} & \underline{3.80} & \underline{6.06} & \underline{4.36} & \underline{5.74} \\
      \textbf{Ours}                          & HRNet-18   & COCO+LVIS & \textbf{1.38} & \textbf{1.46} & \textbf{1.96} & \textbf{3.63} & \textbf{5.83} & \textbf{3.97} & \textbf{5.16} \\
  \bottomrule
  \end{tabular}}
  \end{center}
  \vspace{-0.24cm}
  \caption{Evaluation results on the GrabCut, Berkeley, SBD, and DAVIS datasets.
  The augmented SBD is a combination of the Pascal VOC training set \cite{everingham2010pascal},
  a part of the SBD training set, and a part of the SBD validation set.
  Throughout this essay, the best and the second-best results for different mainstream backbones are written in bold and underlined, respectively.}
  \label{tab:results}
\end{table*}

\begin{table*}[t]
  \begin{center}
  \resizebox{0.99\linewidth}{!}{
  \begin{tabular}{c|l|c|c|c|c|r|c|c|c|c|r}
  \toprule
  \multirow{2}{*}{B} & \multirow{2}{*}{Method} & \multicolumn{5}{c|}{Berkeley} & \multicolumn{5}{c}{DAVIS} \\
      \cline{3-12}
      & & NoF$_{20}$@90\%$\downarrow$ & IoU@5$\uparrow$ & BIoU@5$\uparrow$ & ASSD@5$\downarrow$ & SPC$\downarrow$ & NoF$_{20}$@90\%$\downarrow$ & IoU@5$\uparrow$ & BIoU@5$\uparrow$ & ASSD@5$\downarrow$ & SPC$\downarrow$ \\
      \midrule
      \multirow{6}*{\rotatebox{90}{ResNet-101}} & f-BRS-B \cite{sofiiuk2020fbrs} & 6 & 0.875 & 0.730 & 4.626 & 0.072 & 77 & 0.826 & 0.717 & 11.267 & 0.102 \\
      & FCA-Net \cite{lin2020interactive} & 7 & 0.923 & 0.793 & 2.190 & 0.059 & 74 & 0.867 & 0.771 & 9.048 & \underline{0.075} \\
      & CDNet \cite{lin2020interactive} & \underline{4} & 0.921 & 0.803 & 2.576 & 0.079 & \underline{60} & \underline{0.876} & \underline{0.783} & 8.973 & 0.108 \\
      & FocusCut \cite{lin2022focuscut} & \textbf{3} & \underline{0.933} & \underline{0.811} & \underline{2.050} & 3.152 & 61 & 0.873 & 0.778 & \underline{8.880} & 3.872 \\
      & \textbf{Baseline} & 6 & 0.904 & 0.782 & 4.370 & \textbf{0.052} & 75 & 0.867 & 0.770 & 9.421 & \textbf{0.069} \\ 
      & \textbf{Ours}     & \textbf{3} & \textbf{0.943} & \textbf{0.838} & \textbf{1.789} & \underline{0.057} & \textbf{59} & \textbf{0.893} & \textbf{0.815} & \textbf{8.226} & 0.082 \\ 
      \hline
      \multirow{4}*{\rotatebox{90}{HRNet-18s}} & RITM \cite{sofiiuk2021reviving} \dag & \underline{1} & 0.950 & 0.859 & 1.312 & \textbf{0.035} & \underline{53} & 0.883 & 0.801 & 8.378 & \textbf{0.036} \\
      & FocalClick-S1 \cite{chen2022focalclick} & 4 & 0.946 & 0.849 & 1.565 & 0.065 & 84 & 0.877 & 0.784 & 8.364 & 0.084 \\
      & FocalClick-S2 \cite{chen2022focalclick} & \underline{1} & \underline{0.957} & \textbf{0.889} & \underline{1.170} & 0.052 & 54 & \underline{0.897} & \underline{0.824} & \underline{7.635} & 0.082 \\
      & \textbf{Ours}     & \textbf{0} & \textbf{0.958} & \underline{0.883} & \textbf{1.012} & \underline{0.044} & \textbf{51} & \textbf{0.907} & \textbf{0.830} & \textbf{6.434} & \underline{0.048} \\
  \bottomrule
  \end{tabular}}
  \end{center}
  \vspace{-0.24cm}
  \caption{Comparisons of effectiveness and efficiency on the Berkeley and DAVIS datasets. ``B'' in the table header denotes the term ``backbone''. \dag RITM is the baseline of our method.}
  \label{tab:other_results}
\end{table*}

\begin{figure*}[t]
  \centering
  \includegraphics[width=2.0\columnwidth]{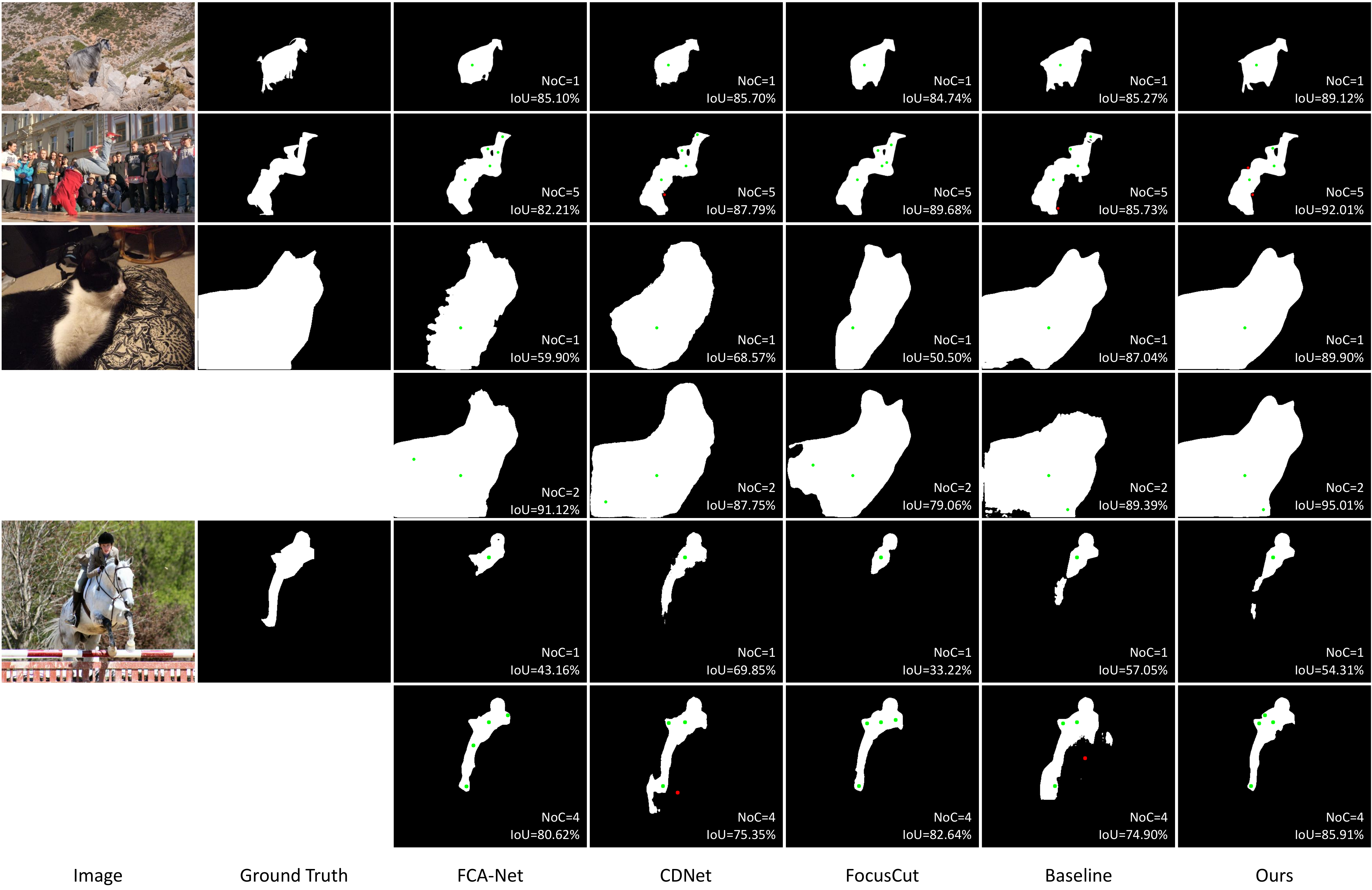}
  \caption{Qualitative comparisons of FCA-Net \cite{lin2020interactive}, CDNet \cite{chen2021conditional}, FocusCut \cite{lin2022focuscut}, our baseline, and our method.
  }
  \label{fig:results}
\end{figure*}

\section{Experiments}
\subsection{Experimental Settings}

\textbf{Segmentation backbones.}
We conducted experiments using
DeepLabV3+ \cite{chen2018encoder} and HRNet+OCR \cite{sun2019high,yuan2020object}
as the baseline network, respectively.
The ResNet-101 \cite{he2016deep}, HRNet-18s \cite{sun2019high}, and HRNet-18 \cite{sun2019high} were employed as backbones.
Improvements introduced by different network baselines are not the focus of this paper.
Therefore, we mainly report experimental results achieved by DeepLabV3+
and provide essential results achieved by HRNet+OCR.
The CFFM was inserted after
the upsampling operator for DeepLabV3+
and the HRNet for HRNet+OCR.
Both the ResNet-101 backbone and the HRNet-18 backbone
were pre-trained on the ImageNet dataset \cite{deng2009imagenet}.

\textbf{Datasets.}
The DeepLabV3+ was trained on the training set of SBD \cite{hariharan2011semantic},
and the HRNet+OCR was trained on the combination of COCO \cite{lin2014microsoft} and LVIS \cite{gupta2019lvis}.
We evaluated our method on four benchmarks:
GrabCut \cite{rother2004grabcut}, Berkeley \cite{martin2001database}, SBD,
and DAVIS \cite{perazzi2016benchmark}.
SBD contains 8,498 images for training,
and its validation set contains 2,820 images including 6,671 instance-level masks.
COCO+LVIS contains 104k images with 1.6M instance-level masks.
GrabCut contains 50 images with an instance mask for each image.
Berkeley consists of 96 images and 100 instance masks.
For DAVIS, we used the same 345 sampled frames
as previous methods \cite{sofiiuk2020fbrs,lin2020interactive,sofiiuk2021reviving,chen2022focalclick,lin2022focuscut}
for evaluation.

\textbf{Implementation details.}
During training,
we resized input images with a random scale factor between 0.75 and 1.40
and randomly cropped them to $320\times 480$.
A horizontal flip and random jittering of brightness, contrast, and RGB values were also applied.
We utilized an Adam optimizer with $\beta_1=0.9$ and $\beta_2=0.999$ to train the network for 60 epochs.
The learning rate for the backbone was initialized to $5\times10^{-5}$ and multiplied by 0.1 on the 50th epoch.
The learning rate for other parts of the segmentation network was 10 times larger than that for the backbone.
The batch size was set to 24.
Following RITM, we used the iterative sampling strategy to simulate user interaction.
For each batch, new annotated clicks were iteratively sampled for 1 to 4 iterations.
In each iteration, up to 24 annotated clicks were randomly sampled
from the eroded mislabelled regions of the last prediction.

During inference, only one click was added in each interaction.
The new click was sampled from misclassified regions
and was the farthest from the region boundaries.
Mathematically, the distance between two points $\bm{p}$ and $\bm{q}$ is defined as
$d(\bm{p},\bm{q})=||\bm{p}-\bm{q}||_2$,
and the distance between a point $\bm{p}$ and a point set $\mathcal P$ is defined as
$d(\bm{p},\mathcal{P})=\text{min}_{\bm{q}\in \mathcal{P}}d(\bm{p},\bm{q})$.
The set of false positive points and false negative points is defined as
$\mathcal{P}_f=\{\bm{p}|\bm{M}_t(\bm{p})=0,\bm{M}_{gt}(\bm{p})=1\}
\cup \{\bm{p}|\bm{M}_t(\bm{p})=1,\bm{M}_{gt}(\bm{p})=0\}$.
For a pixel $p\in{\mathcal{P}_f}$,
the largest connected component in which it is located is denoted as $\mathcal{P}_c(\bm{p})$.
The distance from $\bm{p}$ to the boundaries of $\mathcal{P}_c(\bm{p})$ is defined as
$\eta (\bm{p})=d(\bm{p},\mathcal{P}^C_c(\bm{p}))$,
where $\mathcal{P}^C_c(\bm{p})$ denotes the complement of $\mathcal{P}_c(\bm{p})$.
Following the standard protocol \cite{jang2019interactive,sofiiuk2020fbrs,lin2020interactive,sofiiuk2021reviving,lin2022focuscut,chen2022focalclick},
the new annotated pixel was selected by
$\bm{p}^*=\{\bm{p}|\text{max}_{\bm{p}\in\mathcal{P}_f}\eta (\bm{p})\}$.
The maximum number of clicks was limited to 20 for each sample in all experiments.
Besides, following previous methods \cite{sofiiuk2020fbrs,chen2021conditional,sofiiuk2021reviving,chen2022focalclick},
we adopted test time augmentations,
i.e., the Zoom-In strategy and averaging the predictions of the original image and the horizontally flipped image.

The proposed method was implemented in PyTorch \cite{paszke2019pytorch}.
All the experiments were conducted on a computer with an Intel Xeon Gold 6326 2.90 GHz CPU and NVIDIA GeForce RTX 3090 GPUs.

\textbf{Evaluation metrics.}
The proposed method was evaluated using the following six metrics:
1) NoC@$\alpha$:
the mean number of clicks (NoC) required to reach a predefined intersection over union (IoU) threshold $\alpha$ for all images;
2) IoU@$N$: the mean IoU achieved by a particular NoC $N$;
3) BIoU@$N$: the mean boundary IoU achieved by a particular NoC $N$;
4) ASSD@$N$: the average symmetric surface distance with a particular NoC $N$,
which was used to evaluate the boundary quality of the prediction;
5) NoF$_N$@$\alpha$: the number of failures (NoF) that cannot reach a target IoU $\alpha$ with a certain NoC $N$;
6) SPC: the average running time in seconds per click.

\begin{figure}[t]
  \captionsetup[subfigure]{labelformat=empty,font=small,justification=centering}
  \begin{subfigure}[t]{.5\linewidth}
      \centering
      \includegraphics[width=0.97\columnwidth]{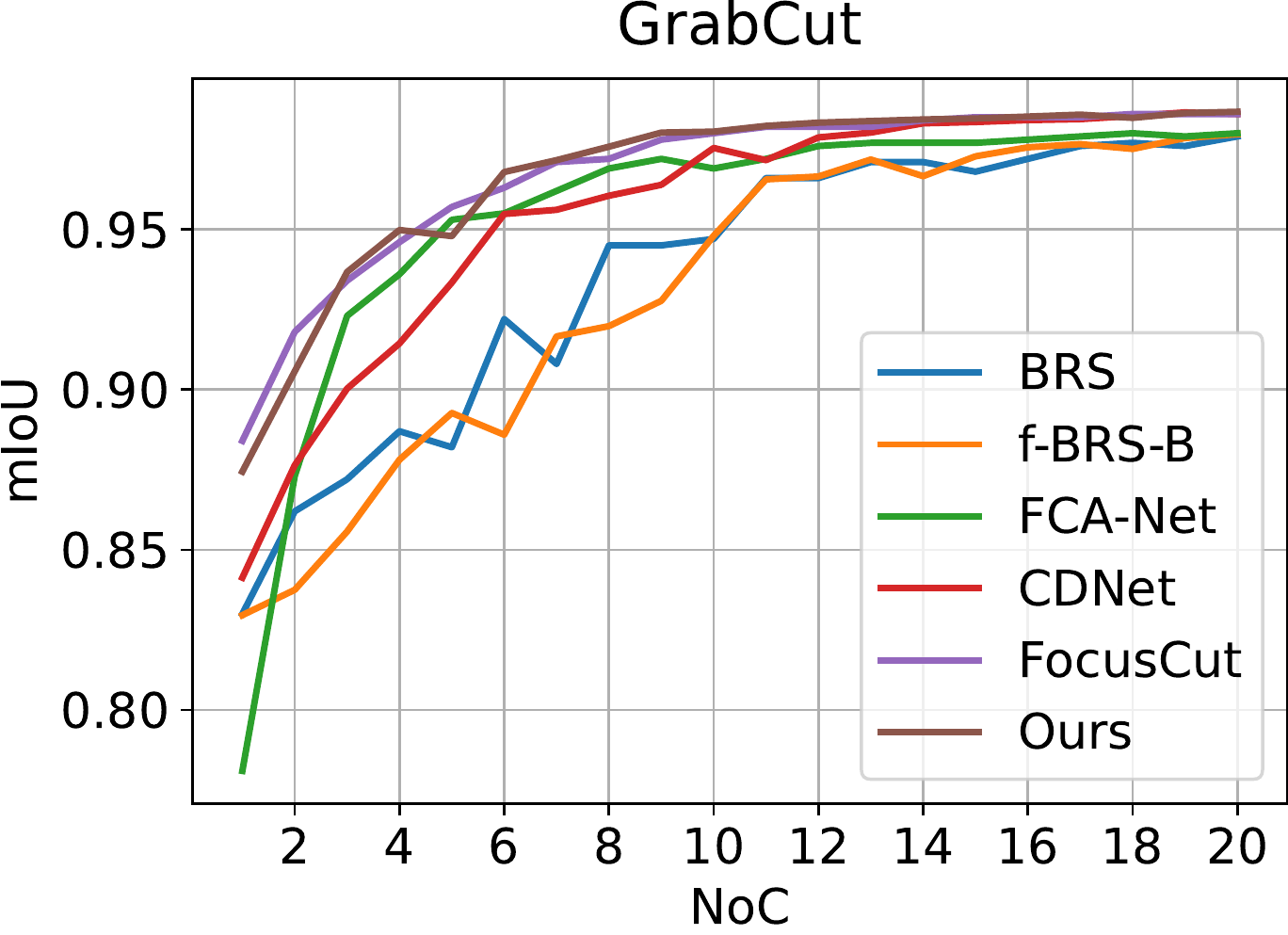}
      \label{fig:chart_grabcut}
  \end{subfigure}%
  \begin{subfigure}[t]{.5\linewidth}
      \centering
      \includegraphics[width=0.97\columnwidth]{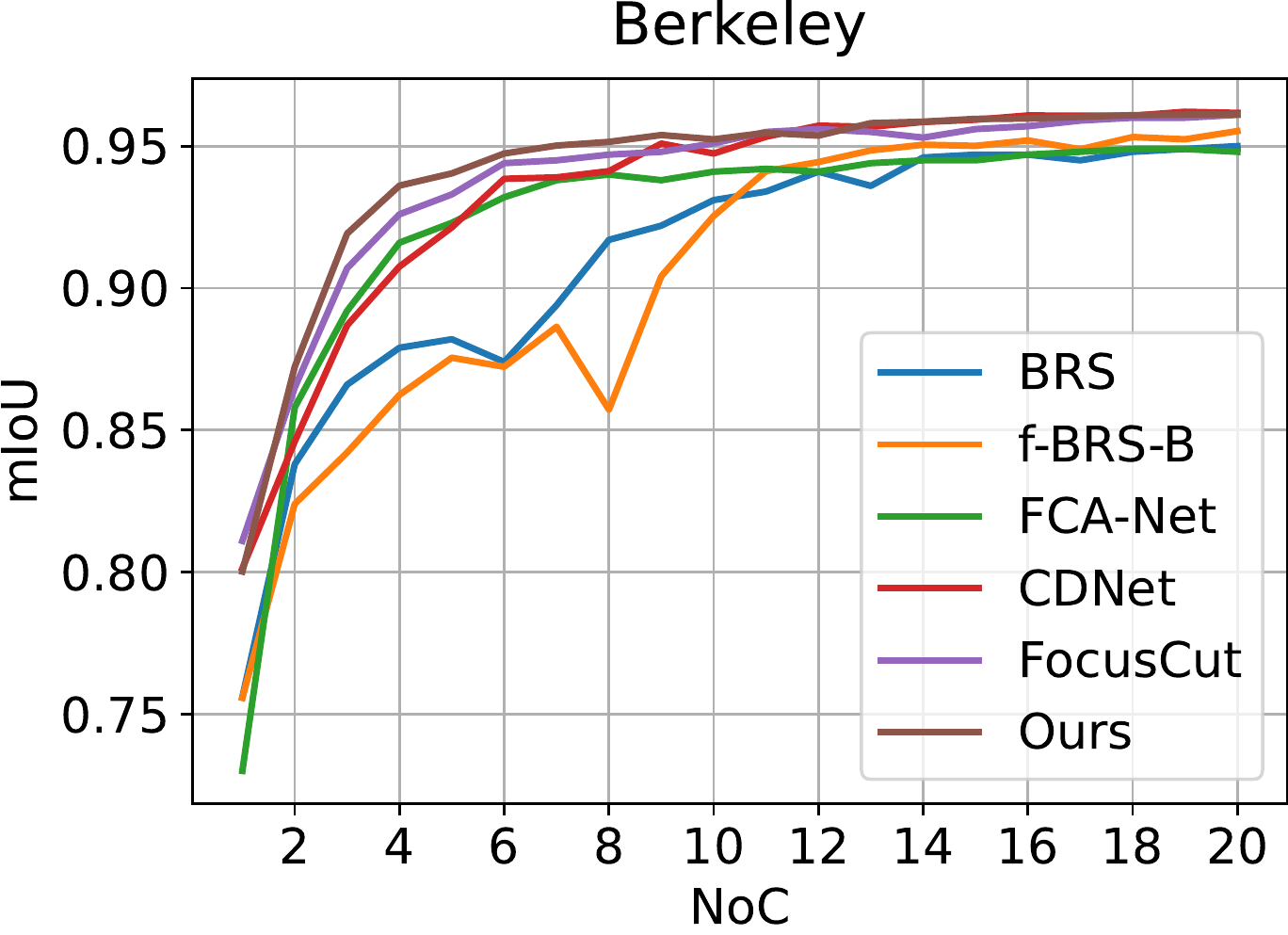}
      \label{fig:chart_berkeley}
  \end{subfigure}%
  \vspace{.1in}
  \begin{subfigure}[t]{.5\linewidth}
      \centering
      \includegraphics[width=0.97\columnwidth]{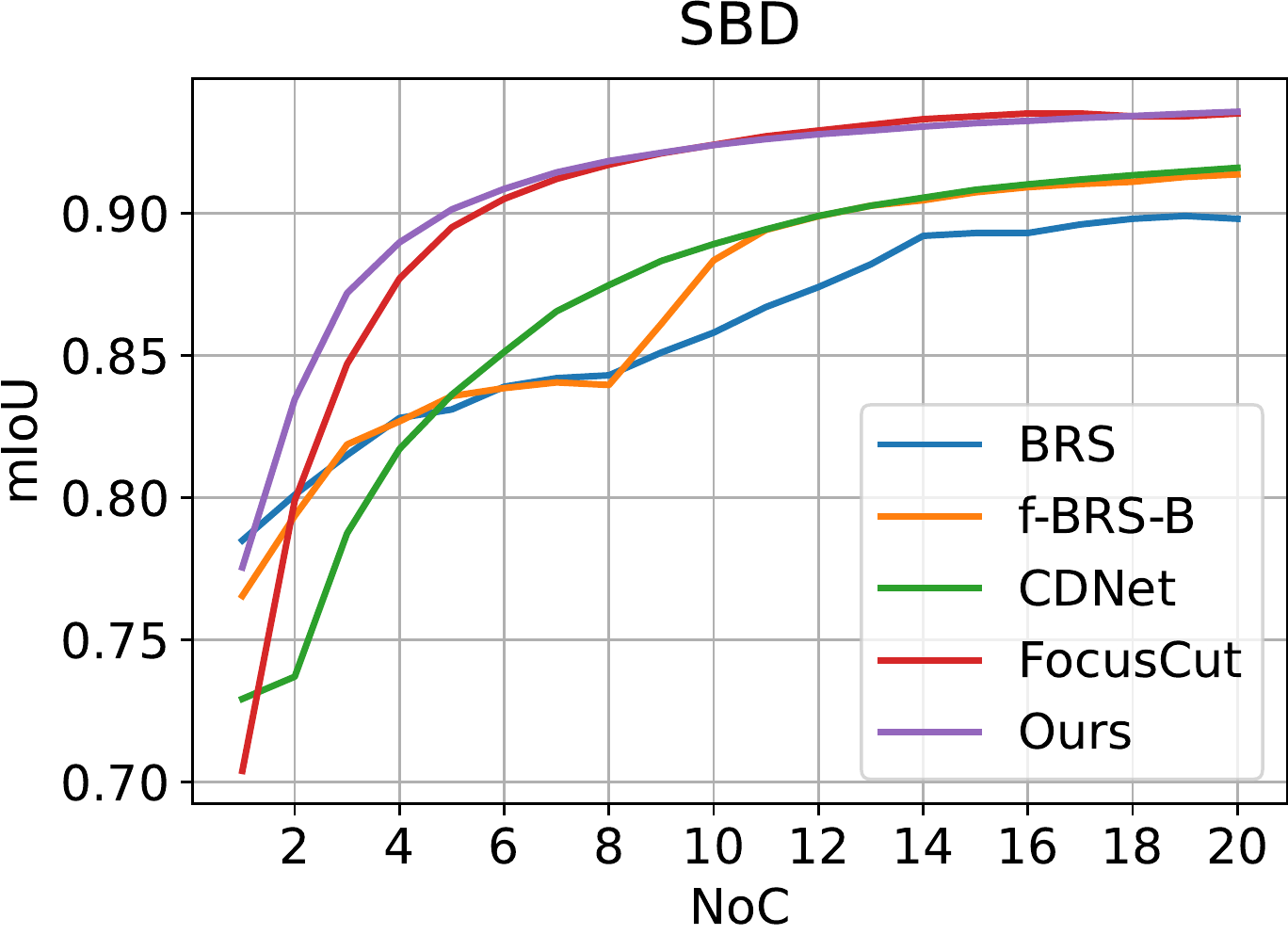}
      \label{fig:chart_sbd}
  \end{subfigure}%
  \begin{subfigure}[t]{.5\linewidth}
      \centering
      \includegraphics[width=0.97\columnwidth]{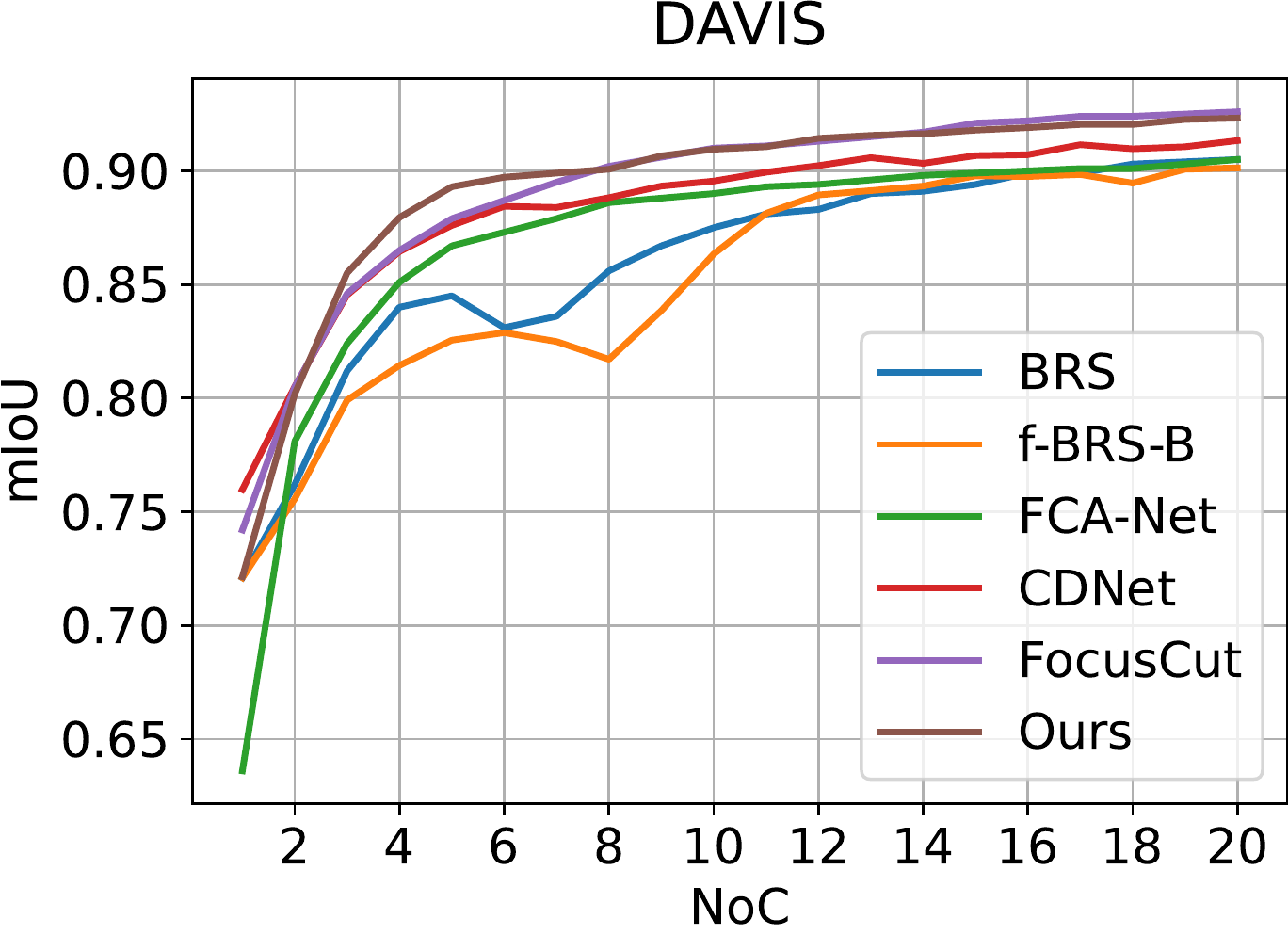}
      \label{fig:chart_davis}
  \end{subfigure}%
  \caption{Comparisons of the mIoU-NoC curves on four datasets.}
  \label{fig:chart}
\end{figure}

\begin{figure}
  \centering
  \includegraphics[width=0.99\columnwidth]{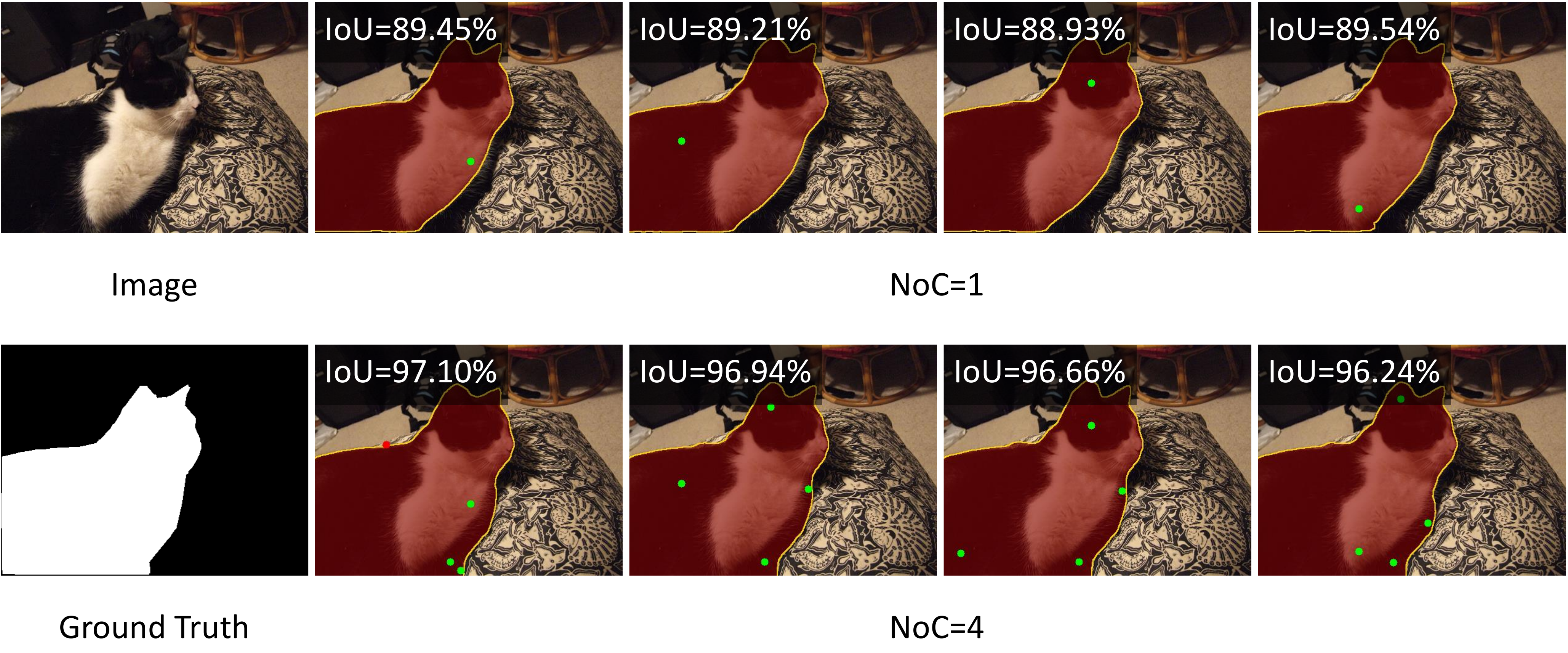}
  \caption{Results obtained by different annotations.}
  \label{fig:click_disturbance}
\end{figure}

\subsection{Comparison with previous work}
\subsubsection{Effectiveness Analysis.}
We have tabulated the quantitative results in Tab. \ref{tab:results} and Tab. \ref{tab:other_results}.
The quantitative results demonstrate that our method can generalize across various datasets and different backbones.
The number of clicks of our method required to reach 85\% and 90\% IoU are much lower than previous methods,
and our method also outperformed previous work in boundary quality.
The results indicate that our method can achieve satisfactory segmentation results with less user effort.
The line charts of mIoU-NoC on four datasets are plotted in Fig. \ref{fig:chart}.
Those results indicate that our method achieved acceptable results given only a few clicks,
steadily improved segmentation results with additional clicks,
and ultimately converged to better results.
Fig. \ref{fig:results} visualizes the qualitative results of some previous work and ours.
Compared with other methods,
our method required fewer clicks to obtain relatively complete segments and fine boundary details.
Additionally, it could handle challenging scenarios,
including color interference (like the goat and the cat),
complex backgrounds (as depicted in the break-dancer  photo), and occlusions (as seen from the racehorse rider).
Please refer to the supplementary material for more visualization examples.
Fig. \ref{fig:click_disturbance} exhibits the sensitivity of our method to click locations.
Our method attained approximate performance for different annotation positions when provided with reasonable annotations.

\subsubsection{Efficiency Analysis.}
Tab. \ref{tab:other_results} presents the average inference speed of different methods,
among which our method achieved a desirable trade-off between speed and accuracy.
Notably, our proposed modules exhibited a low computational budget,
requiring less than 13 ms for all modules,
compared to the baseline.
In summary, the proposed framework achieved competitive results with relatively low computation costs.

\begin{table}[t]
  \begin{center}
  \resizebox{0.99\linewidth}{!}{
  \begin{tabular}{l|l|c|c|c|c}
  \toprule
  \multirow{3}{*}{Method} & \multirow{3}{*}{Backbone} & \multicolumn{2}{c|}{Berkeley} & \multicolumn{2}{c}{DAVIS} \\
  \cline{3-6}
  && NoC    & NoF$_{20}$ & NoC    & NoF$_{20}$ \\
  && @90\%$\downarrow$ & @90\%$\downarrow$ & @90\%$\downarrow$ & @90\%$\downarrow$ \\
  \midrule
  Baseline        & ResNet-101 &  4.31 & 6 & 7.56 & 75 \\ %
  + FFC           & ResNet-101 &  3.75 & 4 & 6.75 & 65 \\ 
  + CFF           & ResNet-101 &  3.07 & \textbf{1} & 6.62 & 66 \\ 
  + FFC + CFF     & ResNet-101 &  \textbf{2.84} & 3 & \textbf{6.48} & \textbf{59} \\
  \hline
  Baseline    & HRNet-18 &  2.60 & 1 & 5.73 & 54 \\
  + FFC       & HRNet-18 &  2.11 & 1 & 5.52 & 54 \\
  + CFF       & HRNet-18 &  2.05 & \textbf{0} & 5.32 & 52 \\
  + FFC + CFF & HRNet-18 &  \textbf{1.96} & \textbf{0} & \textbf{5.16} & \textbf{51} \\
\bottomrule
\end{tabular}}
\end{center}
\vspace{-0.24cm}
\caption{An ablation study for the core components.}
\label{tab:ablation}
\end{table}


\begin{figure*}[t]
  \centering
  \includegraphics[width=2.0\columnwidth]{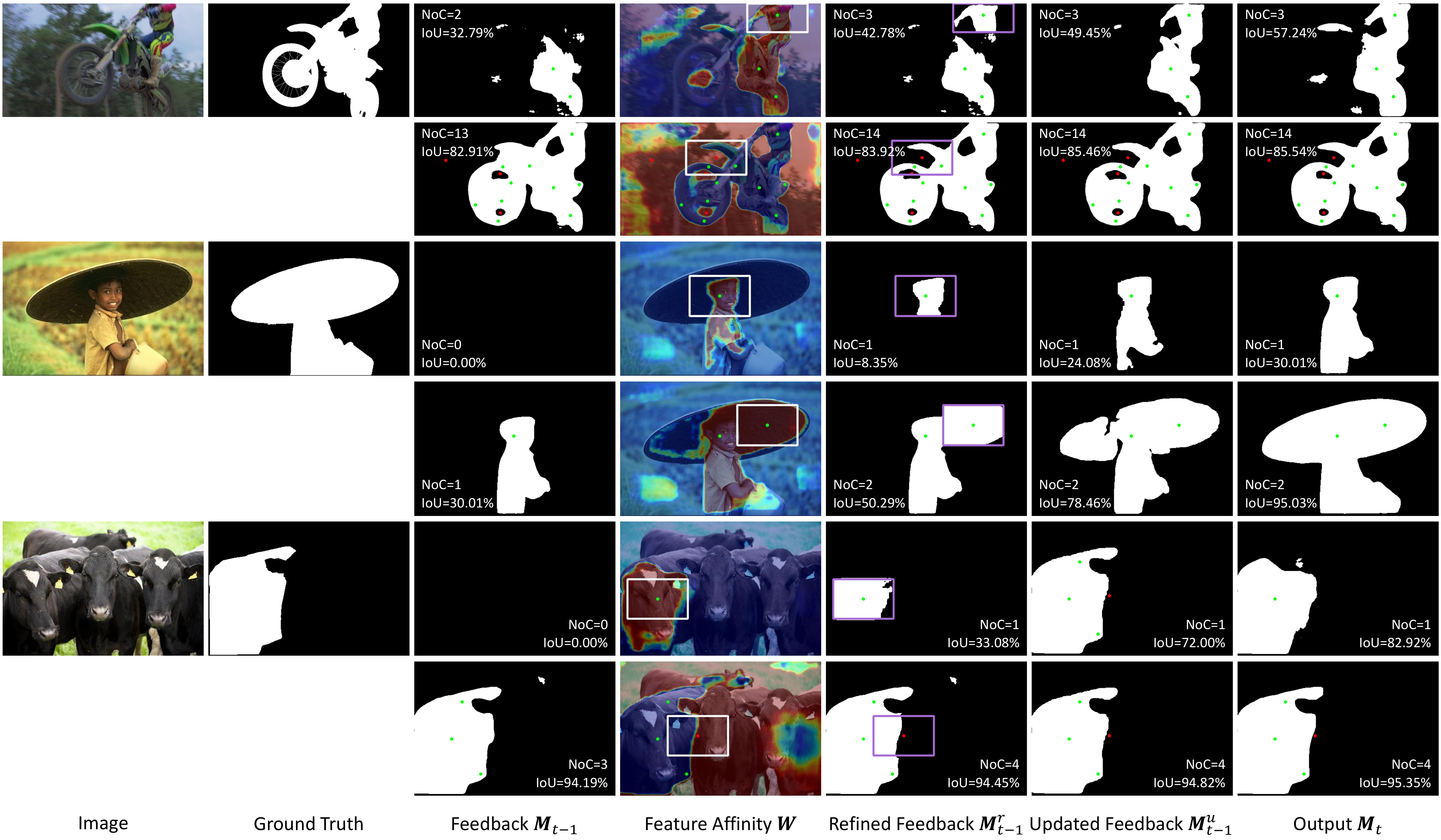}
  \caption{Visualization of the feedback, the feature affinity, the refined feedback, the updated feedback, and the output.
  The feature affinity is shown in a heatmap,
  where red color denotes high feature affinity, and blue color denotes low feature affinity.
  }
  \label{fig:ablation_restuls}
\end{figure*}

\begin{table*}[t]
  \begin{minipage}{0.3\linewidth}
    \begin{center}
      \resizebox{0.99\linewidth}{!}{
      \begin{tabular}{l|c|c|c|c}
        \toprule
        \multirow{2}{*}{\shortstack{Similarity\\Measurement}} & \multicolumn{2}{c|}{Berkeley} & \multicolumn{2}{c}{DAVIS} \\
        \cline{2-5}
        & @85\% & @90\% & @85\% & @90\% \\
        \midrule
        Exponential & 2.06 & 3.60 & 5.26 & 7.51 \\
        \textbf{Cosine} & \textbf{1.88} & \textbf{2.85} & \textbf{4.60} & \textbf{5.82} \\
        \bottomrule
      \end{tabular}}
  \end{center}
  \vspace{-0.24cm}
  \caption{The mean NoC with respect to the similarity measurement.
  The exponential similarity is defined as $\bm{W}(i,j)=e^{-||\bm{F}(i,j)-\bm{F}(x_{new},y_{new})||^2/\sigma}$.}
  \label{tab:similarity_measurement_for_ffcm}
\end{minipage}
  \hfill
  \begin{minipage}{0.285\linewidth}
    \begin{center}
      \resizebox{0.99\linewidth}{!}{
      \begin{tabular}{c|c|c|c|c}
        \toprule
        \multirow{2}{*}{\shortstack{Expansion\\Ratio $r$}} & \multicolumn{2}{c|}{Berkeley} & \multicolumn{2}{c}{DAVIS} \\
        \cline{2-5}
        & @85\% & @90\% & @85\% & @90\% \\
        \midrule
        0.1      & 1.91 & 3.43 & 5.52 & 7.46 \\ 
        0.2      & 1.97 & 3.10 & 5.12 & 6.82 \\
        \textbf{0.3} & 1.88 & \textbf{2.85} & \textbf{4.60} & \textbf{5.82} \\
        0.4      & \textbf{1.79} & 3.12 & 4.74 & 6.42 \\
        0.5      & 1.84 & 3.27 & 5.15 & 6.72  \\
        \bottomrule
      \end{tabular}}
  \end{center}
  \vspace{-0.24cm}
  \caption{The mean NoC with respect to the expansion ratio $r$.}
  \label{tab:expansion_ratio}
  \end{minipage}
  \hfill
  \begin{minipage}{0.393\linewidth}
    \begin{center}
      \resizebox{0.99\linewidth}{!}{
      \begin{tabular}{l|c|c|c|c}
        \toprule
        \multirow{2}{*}{Method} & \multicolumn{2}{c|}{Berkeley} & \multicolumn{2}{c}{DAVIS} \\
        \cline{2-5}
        & @85\% & @90\% & @85\% & @90\% \\
        \midrule
        w/o feedback                 & 2.31 & 4.31 & 5.23 & 7.56 \\
        Concat feedback with input   & 2.01 & 4.00 & 5.03 & 7.08  \\ 
        CFFM w/o residual connection & 1.98 & 3.52 & 4.91 & 6.63 \\
        CFFM w/o gate                & 1.96 & 3.36 & 4.85 & 6.37  \\
        \textbf{CFFM}                & \textbf{1.81} & \textbf{3.07} & \textbf{4.75} & \textbf{6.10}  \\
        \bottomrule
      \end{tabular}}
    \end{center}
    \vspace{-0.24cm}
    \caption{The mean NoC with respect to different feedback fusion architectures.}
    \label{tab:architecture_for_cffm}
  \end{minipage}
\end{table*}

\subsection{Ablation Studies}
\label{sec:ablation}
We evaluated the efficacy of each proposed component.
Berkeley and DAVIS were chosen as the main evaluation datasets
because they cover challenging scenarios, such as unseen categories, motion blur, and occlusions.
Besides, they have better annotation quality than SBD.

Tab. \ref{tab:ablation} tabulates the quantitative ablation results.
To integrate the feedback into the network,
the corrected feedback $\bm{M}^c_{t-1}$ was concatenated with the features $\bm{F}$
in the ``+ FFC'' variant.
When only one module was embedded,
both the FFCM and the CFFM improved the results.
The CFFM boosted the performance more;
this proves effectiveness of deep feedback integration.
The FFCM has also improved the results.
The reason is that the CFFM relies on the quality of the feedback,
and the FFCM provides refined feedback for it.
With the two modules working synergistically,
our method significantly reduced the NoC and NoF.

Qualitative results for the FFCM and the CFFM are shown in Fig. \ref{fig:ablation_restuls}.
The FFCM utilizes the feature affinity to refine the feedback from a local perspective.
For instance, in the 14th round of interaction,
the FFCM yielded finer segmentation boundaries on the front fender of the motorcycle.
The CFFM refines the feedback in a global view
and updates the deep features to improve the overall segmentation results,
e.g., the boy wearing a hat and the bull.

\textbf{Analysis for the FFCM.}
In Tab. \ref{tab:similarity_measurement_for_ffcm},
we analyzed the effect of different similarity measurements,
e.g. exponential similarity and cosine similarity.
Using cosine similarity to measure the feature affinity achieved better results.
We also verified the robustness of the FFCM with respect to the expansion ratio $r$.
As illustrated in Tab. \ref{tab:expansion_ratio},
an expansion ratio of 0.3 yielded the best results.

\textbf{Analysis for the CFFM.}
Different implementations of feedback fusion have been explored:
1) directly concatenating feedback with the input,
2) fusing feedback into deep features but removing the residual connection,
i.e., removing the second term of Eq. \ref{eq:update_feature},
and 3) fusing feedback into deep features but removing the gate.
As shown in Tab. \ref{tab:architecture_for_cffm},
integrating feedback into the network improves the performance,
which demonstrates the instructive ability of the feedback.
Compared with directly concatenating feedback with the input,
feedback fusion in deep layers significantly reduces the required number of annotated clicks
by 1.24 NoC@90\% on Berkeley and 1.56 NoC@90\% on DAVIS.

\subsection{Limitations}
Although our method benefits from the feedback guidance,
it still has certain limitations.
First,
there is no guarantee that each round of interaction yields superior results compared to the previous one.
Second, ambiguity has yet to be resolved in our method.
For example, if a click is provided on a shirt,
both the shirt and the person wearing it are likely to be the target object.
Additionally, our method may struggle when handling thin structures,
such as ropes, insect legs, and bicycle spokes.

\section{Conclusion}
The segmentation result from the last interaction (feedback) provides instructive information about the target object to the current interaction.
To exploit the feedback,
this paper proposes a deep feedback integration approach called FCFI.
FCFI first performs local refinement on the feedback.
Then, it collaboratively and globally updates the feedback and the features in deep layers of the segmentation network.
We have experimentally demonstrated that FCFI has strong generalization capabilities
and outperformed many previous methods with fast processing speed.

\section*{Acknowledgment}
This research was supported by the National Key R\&D Program of China (2020YFB1708900) and the National Nature Science Foundation of China (62021002).

{\small
\bibliographystyle{ieee_fullname}
\bibliography{egbib}
}

\end{document}